# Domain-Specific Face Synthesis for Video Face Recognition From a Single Sample Per Person

Fania Mokhayeri ⬤, *Student Member, IEEE*, Eric Granger ⬤, *Member, IEEE*, and Guillaume-Alexandre Bilodeau ⬤, *Member, IEEE*

*Abstract*—In video surveillance, face recognition (FR) systems are employed to detect individuals of interest appearing over a distributed network of cameras. The performance of still-to-video FR systems can decline significantly because faces captured in unconstrained operational domain (OD) over multiple video cameras have a different underlying data distribution compared to faces captured under controlled conditions in the enrollment domain with a still camera. This is particularly true when individuals are enrolled to the system using a single reference still. To improve the robustness of these systems, it is possible to augment the reference set by generating synthetic faces based on the original still. However, without the knowledge of the OD, many synthetic images must be generated to account for all possible capture conditions. FR systems may, therefore, require complex implementations and yield lower accuracy when training on many less relevant images. This paper introduces an algorithm for domain-specific face synthesis (DSFS) that exploits the representative intra-class variation information available from the OD. Prior to operation (during camera calibration), a compact set of faces from unknown persons appearing in the OD is selected through affinity propagation clustering in the captured condition space (defined by pose and illumination estimation). The domain-specific variations of these face images are then projected onto the reference still of each individual by integrating an image-based face relighting technique inside the 3-D reconstruction framework. A compact set of synthetic faces is generated that resemble individuals of interest under the capture conditions relevant to the OD. In a particular implementation based on sparse representation classification, the synthetic faces generated with the DSFS are employed to form a cross-domain dictionary that accounts for structured sparsity, where the dictionary blocks combine the original and synthetic faces of each individual. Experimental results obtained with videos from the Chokepoint and COX-S2V data sets reveal that augmenting the reference gallery set of still-to-video FR systems using the proposed DSFS approach can provide a significantly higher level of accuracy compared with the state-of-the-art approaches, with only a moderate increase in its computational complexity.

Manuscript received January 13, 2018; revised May 15, 2018 and July 18, 2018; accepted July 21, 2018. Date of publication August 21, 2018; date of current version September 13, 2018. This work was supported by NSERC Grant. The associate editor coordinating the review of this manuscript and approving it for publication was Dr. Adams W. K. Kong. *(Corresponding author: Fania Mokhayeri.)*

F. Mokhayeri and E. Granger are with the Laboratory for Imagery, Vision and Artificial Intelligence, École de technologie supérieure, Université du Québec, Montreal, QC H3C 1K3, Canada (e-mail: fmokhayeri@livia.etsmtl.ca; eric.granger@etsmtl.ca).

G.-A. Bilodeau is with the Image and Video Processing Laboratory, Polytechnique Montréal, Montreal, QC H3T 1J4, Canada (e-mail: gabilodeau@polymtl.ca).

Color versions of one or more of the figures in this paper are available online at http://ieeexplore.ieee.org.

Digital Object Identifier 10.1109/TIFS.2018.2866295

## I. INTRODUCTION

STILL-TO-VIDEO face recognition (FR) is an important function in several video surveillance applications, particularly for watch-list screening. Given one or more reference still images of a target individual of interest, still-to-video FR systems seeks to accurately detect their presence in videos captured over multiple distributed surveillance cameras [1].

Despite the recent progress in computer vision and machine learning, designing a robust system for still-to-video FR remains a challenging problem in real-world surveillance applications. One key issue is the visual domain shift between faces from the enrollment domain (ED), where reference still images are typically captured under controlled conditions, and those from the operational domain (OD), where video frames are captured under uncontrolled conditions with variations in pose, illumination, blurriness, etc. The appearance of faces captured in videos corresponds to multiple non-stationary data distributions that can differ considerably from faces captured during enrollment [2]. Another key issue is the limited number of reference stills that are available per target individual to design facial models. Although still faces from the cohort or other non-target persons, and trajectories of video frames from unknown individuals are typically available. In many surveillance applications (e.g., watch-list screening), only a single reference still per person is available for design, which corresponds to the so-called Single Sample Per Person (SSPP) problem. The performance of still-to-video FR systems can decline significantly due to the limited information available to represent the intra-class variations seen in video frames. Many discriminant subspaces and manifold learning algorithms cannot be directly employed with a SSPP problem. It is also difficult to apply representation-based FR methods such as sparse representation-based classification (SRC) [3].

Different techniques for SSPP problems have been proposed to improve the robustness of FR systems, such as using multiple face representations [2], face frontalization [4], generating synthetic faces from the original reference stills [5], [6], synthesis of face model parameters [7], and incorporating generic auxiliary set[1] [8], [9]. This paper focuses on methods that are based on augmenting the reference gallery set through synthetic set generated based on the original reference still, and by taking into account the intra-class variation information transferred from a generic set. A challenge with strategies for augmenting the reference gallery set is selecting a sufficient number of synthetic or generic faces to cover intra-class variations in the OD. Many synthetic faces or generic auxiliary faces may be generated or collected, respectively, to account for all possible capture conditions. In this case, FR systems would, therefore, require complex implementations and may yield lower accuracy when training on many facial images that provide less relevant information for FR in the OD. Another challenge is domain discrepancy between synthetic and real images. The synthetically generated images may not be covering the range intra-class variations of OD, since they are highly correlated with the original face images.

---

[1]A generic set is defined as an auxiliary set that contains multiple video frames per person from other unknown individuals. It provides an abundance of information on intra-class variations of the capture conditions.

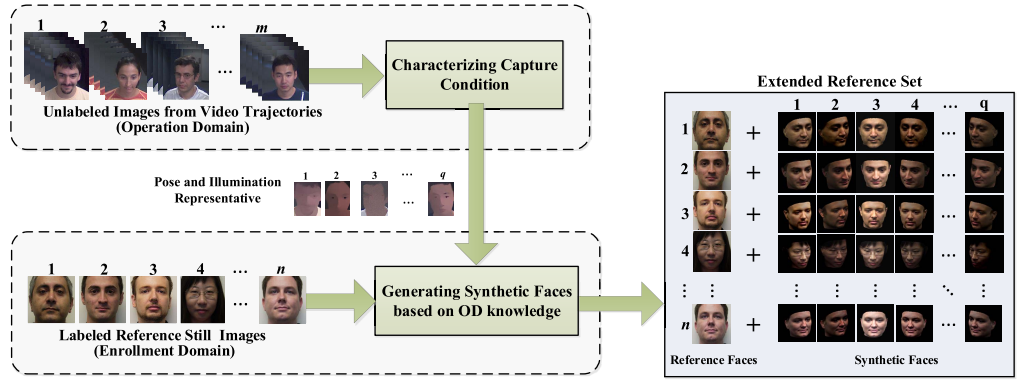

Fig. 1. Overview of the proposed DSFS algorithm to augment the reference gallery set. We assume that the gallery set initially contains only one reference still image per individual of interest.

In this paper, a new approach is proposed that exploits the discriminant information of the generic set for the face synthesis process. The new algorithm called domain-specific face synthesis (DSFS) maps representative variation information from the generic set in the OD to the original reference stills. In this way, a compact set of synthetic faces is generated that represent reference still images and probe video frames under a common capture condition. As depicted in Fig. 1, the DSFS technique involves two main steps: (1) characterizing capture condition information from the OD, (2) generating synthetic face images based on the information obtained in the first step. Prior to operation (during camera calibration process), a generic set is collected from video captured in the OD. A compact and representative subset of face images is selected by clustering this generic set in a capture condition space defined by pose, illumination, blur. The 3D model of each reference still image is reconstructed via a 3D morphable model and rendered based on pose representatives. Finally, the illumination-dependent layers of the lighting representatives are extracted and projected on the rendered reference images with the same pose. In this manner, domain-specific variations are effectively transferred onto the reference still images. The major contributions of our work are:

- A technique based on affinity clustering to select representative facial exemplars using information extracted from videos captured form operational domain. This prevents over-fitting of classifier due to the redundant information and improves efficiency.

- A novel face synthesizing technique that maps the intra-class variation from facial exemplars available in the operational domain to generate a representative set of face images under real-world capture conditions.
- A technique to design a compact and discriminative dictionary for SRC allowing to perform robust still-to-video FR with only one reference still ROI.

In a particular implementation for still-to-video FR, the original and synthetic face images are employed to design a structural dictionary with powerful variation representation ability for SRC. The dictionary blocks represent intra-class variations computed from either the reference faces themselves or the synthetic faces [10]. The cooperation of SRC with the proposed DSFS improves the robustness of SRC for video-based FR in a SSPP scenario to domain variations. In order to validate the performance of the proposed DSFS algorithm for still-to-video FR with a SSPP, this SRC implementation is evaluated and compared on two public face databases.

The main advantage of the proposed approach is the ability to provide a compact set that can accurately represent the original reference face with relevant of intra-class variations in pose, illumination, motion blur, etc., corresponding to capture condition in the OD. For instance, in the context of SRC implementations, this set can prevent over-fitting and refines more informative classes during the sparse coding process.

The rest of the paper is organized as follows. Section II provides an overview of related works for FR with a SSPP. Section III describes the proposed face synthesizing algorithm. Section IV presents a particular implementation of the DSFS for still-to-video FR system. In Section V, the experimental methodology (dataset, protocol, and performance metrics) for validation of FR systems is described and the experimental results is presented. Finally, Section VII concludes the paper and discusses some future research directions.

## II. RELATED WORK – STILL-TO-VIDEO FACE RECOGNITION FROM A SINGLE STILL

Several techniques have been proposed in the literature to improve the robustness of still-to-video FR systems designed using a SSPP. They can be categorized into techniques for (1) multiple face representation, (2) generic learning, and (3) generation of synthetic faces. An overview of the techniques is presented as bellow.

### A. Multiple Face Representations

One effective approach to address the SSPP problem in FR is to extract discriminant features from face images. Bashbaghi *et al.* [2], [11] proposed a robust still-to-video FR system based on diverse face representations. They applied multiple appearance-invariant feature extraction techniques to patches isolated from the reference still images in order to produce multiple face representations and generate a pool of diverse exemplar-SVMs. This pool provides robustness to common nuisance factors encountered in surveillance applications. Lu *et al.* [12] proposed a discriminative multi-manifold analysis method by learning discriminative features from image patches. In this technique, the patches of each individual are considered to form a manifold for each sample per person and a projection matrix is learned by maximizing the manifold margin of different persons. In [13], a face image is processed by several pose-specific deep convolutional neural network (CNN) models to generate multiple pose-specific features. The multiple face representation techniques are, however, able to compensate only the small variations and consequently are not effective to tackle with variations in practical applications (e.g., extreme illumination, pose and expression variations).

### B. Generic Learning

An early finding to compensate visual domain shift in FR systems is to employ a generic set to enrich the diversity of the reference gallery set that is the so-called generic learning concept [14]. Generic learning has been widely discussed by many researchers [15], [16]. Su *et al.* proposed an adaptive generic learning method for FR which utilized external data to estimate the within-class scatter matrix for each individual and applies this information to the reference set [14]. In recent years, integration of sparse representation-based classification (SRC) with generic learning for FR has attracted significant attention. Deng *et al.* [8] added generic leaning into the SRC framework and proposed the extended SRC (ESRC), which provide additional information from other face datasets to construct an intra-class variation dictionary to represent the changes between the training and probe images. Yang *et al.* [17] introduced a sparse variation dictionary learning (SVDL) technique by taking the relationship between the reference set and the external generic set into account and obtained a projection by learning from both generic and reference set. In [18], intra-class variation information from the OD is integrated with the reference set through domain adaptation to enhance the facial models. Authors in [19] proposed a robust auxiliary dictionary learning (RADL) technique that extracts representative information from generic dataset via dictionary learning without assuming prior knowledge of occlusion in probe images. Zhu *et al.* proposed a local generic representation-based framework (LGR) for FR with SSPP [20]. It builds a gallery dictionary by extracting the neighboring patches from the gallery dataset, while an intra-class variation dictionary is constructed by using an external generic training dataset to predict the intra-class variations. Authors in [21] proposed a robust still-to-video FR using a multi-classifier system in which each classifier is trained by a reference face still versus many lower-quality faces of non-target individuals captured in videos. In this system, the auxiliary set collected from videos of unknown people in the OD are employed to select discriminant feature sets and ensemble fusion functions. In [22], a supervised autoencoder network is proposed for still-to-video FR system to generate canonical face representations from unknown video frames in the OD that are robust to appearance variations. Despite the significant improvements reported with generic learning, several critical issues remain to be addressed. The generic intra-class variation may not be similar to that of gallery individuals, so the extraction of discriminative information from the generic set may not be guaranteed. Moreover, the large number of images collected from external data may contain redundant information which could lead to complex implementations and degrade the capability in covering intra-class variations.

### C. Synthetic Face Generation

Augmenting the reference gallery set synthetically is another strategy to compensate the appearance variations in FR with SSPP. Shao *et al.* in [23] presented a SRC-based FR algorithm that extends the dictionary using a set of synthetic faces generated by calculating the image difference of a pair of faces. Authors in [24] augmented the reference gallery set by generating a set of synthetic face images under camera-specific lighting conditions to design a robust still-to-video FR system under surveillance conditions. 3D Morphable Model (3DMM), proposed by Blanz and Vetter [25], has been widely used to synthesize new face images from a single 2D face image. In the past decade, several extension of this technique is presented. Authors in [26] employed a CNN to regress 3DMM shape and texture parameters directly from an input image without an optimization process which renders the face and compares it to the image. Zhang *et al.* proposed a 3D Spherical Harmonic Basis Morphable Model (SHBMM) that is an integration of spherical harmonics into the 3DMM framework [27]. Richardson *et al.* proposed a neural network for reconstructing a detailed facial surface in 3D from a single image where the rough facial geometries are modeled using a 3DMM and facial features of that geometry is refined by a CNN [28]. The CNN proposed in [29] integrates an expert-designed decode layer that implements an elaborate generative analytically-differentiable image formation model on the basis of a detailed parametric 3D face model. Apart from 3D reconstruction techniques, some 2D-based techniques generate synthetic images under various illumination conditions by transferring the illumination of target images to the reference face images [24], [30]. Recently, generative adversarial network (GAN), introduced by Goodfellow et al. [31], has become popular for realistic face synthesis [32]–[34]. These methods formulate GAN as a minimax game, where a discriminator distinguishes face images in the real and synthetic domains, while a generator reduces its discriminativeness by synthesizing realistic face images. The competition converges when the discriminator is unable to differentiate between real and synthetic domains. Shrivastava *et al.* in [35] proposed Simulated+Unsupervised learning method that improves the realism of synthetic images. The proposed learning method employs an adversarial network similar to GAN with synthetic images as inputs instead of random vectors. Although synthetic images can improve the robustness of FR systems designed with a SSPP, they may not be covering the range intra-class variations in practical scenarios because of redundancy in the learned discriminative subspace. Many synthetic images should be generated to account for all possible capture conditions in ODs. Without the selection of representative face images from both the reference gallery and external data, generating the synthetic faces may require complex implementations and yield lower accuracy when training on many less relevant images.

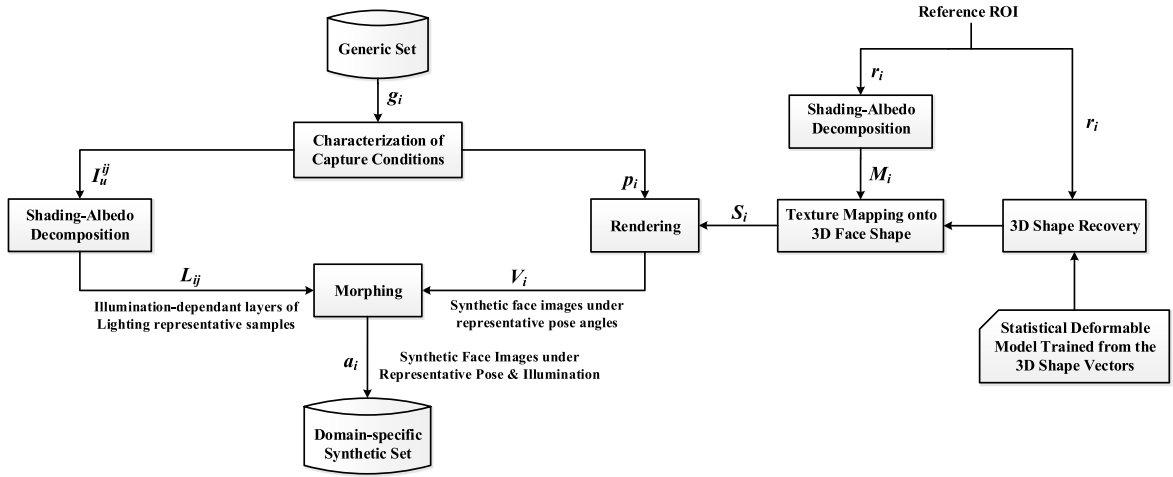

Fig. 2. Block diagram of the DSFS technique applied to a reference still ROI.

### D. Synthesis of Face Model Parameters

Another effective solution to address the appearance variations in FR with SSPP is synthesis of the face model parameters or representations to directly train recognition algorithm from synthesis parameters. Sanderson *et al.* proposed a statistical framework to extend each frontal face model to artificially synthesized models for non-frontal views based on maximum likelihood linear regression and standard multivariate linear regression methods [7]. In [36], a novel face classification approach based on Active Appearance Model is proposed, where pose-robust features are obtained without the image synthesis step. In [25], images are analyzed by fitting a statistical model of 3D faces to images and recognition is performed based on model coefficients, which represent intrinsic shape and texture of faces, and are independent of the imaging conditions. In [37], an approach for synthesized intermediate representations (e.g., frontal to profile face images) is proposed which gradually reduces the reconstruction residue of the target data to link the ED and OD. Although using model parameters directly for classification allows skipping the artifact producing image synthesis step, it is not able to convey the full range of real-world intra-class variations.

To overcome the challenges discussed above, this paper presents a framework that exploits both face synthesis and generic learning. The technique proposed in Section III generates a compact set of synthetic facial images per individual of interest that corresponds to relevant OD capture conditions, by mapping the intra-class variations from a representative set of video frames selected from the OD into the original reference still images.

## III. DOMAIN-SPECIFIC FACE SYNTHESIS

This paper focuses on augmenting reference face set to cover the intra-class variations of individual appearing in ODs with a compact set of synthetic face images. A new Domain-Specific Face Synthesis (DSFS) technique is proposed that employs knowledge of the OD to generate a compact set of synthetic face images for the design of FR systems.

Prior to operation, e.g., during a camera calibration process, DSFS selects facial regions of interest (ROIs) isolated in videos with representative pose angles and illumination conditions from facial trajectories of unknown persons captured in the OD. These video ROIs are selected via clustering facial trajectories in the captured condition space defined by pose and illumination conditions. Next, the DSFS exploits a 3D shape reconstruction method and an image-based illumination transferring technique to generate synthetic ROIs under representative pose angles and illumination conditions from the reference still ROIs. To do so, the 3D models of the reference still ROIs are reconstructed and rendered w.r.t. the representative pose angles. The illumination-dependent layers of the representative illumination conditions are then extracted and projected onto rendered images with the same view by applying a morphing between the layers. In other words, illumination-dependent layers of video ROIs from the OD are replaced with that of the still reference ROI from the ED. Fig. 2 shows the pipeline of the DSFS technique.

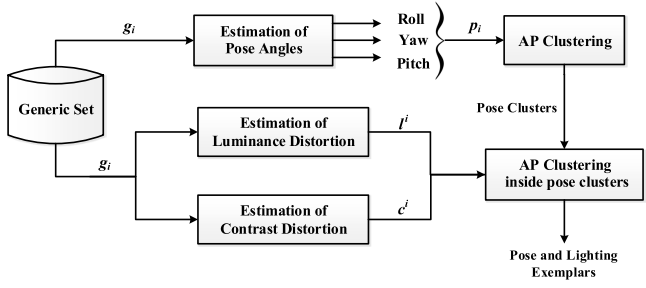

Fig. 3. Pipeline for characterizing capture conditions of video ROIs in the operational domain.

## A. Characterizing the Capture Conditions

An important concern for the reference set augmentation is the selection of representative pose angles and illumination conditions to represent relevant capture conditions in the OD. As mentioned, adding a large number of potentially redundant images to the reference set can significantly increase the time and memory complexity, and may degrade the recognition performance due to over-fitting.

With the DSFS technique, the representative pose angles and illumination conditions to cover relevant intra-class variations is approximated by characterizing the capture conditions from a large generic set of video ROIs. This set is formed with multiple ROIs isolated in several facial trajectories of unknown persons captured in the OD. Let $\mathbf{R} = \{\mathbf{r}_i \in \mathbb{R}^{d \times d} | i = 1, \ldots, n\}$ be a set of ROIs of still reference individuals, and $\mathbf{G} = \{\mathbf{g}_i \in \mathbb{R}^{d \times d} | i = 1, \ldots, m\}$ be a video ROIs in the generic set, where $n$ and $m$ denote the number of individuals in the reference gallery set, and the generic set, respectively.

In the proposed technique (see Fig. 3), an estimation of luminance, contrast and pose are measured for each video ROIs in the generic set $\mathbf{g}_i$. Next, a two-step clustering process is applied on video ROIs in the measurement space defined by pose, luminance and contrast. The first step is applied on all ROIs in the 3D metric space defined by pose (tilt, yaw and roll), while the second step is applied on ROIs of each pose cluster in the 2D space defined by luminance and contrast metrics. The prototype of each cluster is considered as an exemplar. The generic variational information obtained during this step is then transferred to the reference still ROIs during the face synthesizing step (see Section III B). Although many algorithms are also suitable to implement DSFS, the following subsections describe DSFS with specific algorithms.

### 1) Estimation of Head Pose:
The estimate of head pose for the $i^{\text{th}}$ video ROI ($\mathbf{g}_i$) in the generic set is defined as $\mathbf{p}_i = (\theta_i^{pitch}, \theta_i^{yaw}, \theta_i^{roll})$. Euler angles $\theta_i^{pitch}$, $\theta_i^{yaw}$, and $\theta_i^{roll}$ are used to represent pitch, yaw and roll rotation around $X$ axis, $Y$ axis, and $Z$ axis of the global coordinate system, respectively. In order to estimate the head pose, the discriminative response map fitting (DRMF) method [38] is employed. It is the current state-of-the-art method in terms of fitting accuracy and efficiency suitable for handling occlusions and changing illumination conditions.

### 2) Luminance-Contrast Distortion:
Luminance and contrast distortion measures estimate the distortion between a video ROI and the corresponding reference still ROI. Components of the structural similarity index measure presented [39] are employed to measure the proximity of the average luminance and contrast locally by utilizing sliding window. The global luminance distortion in image quality (GLQ) factor between $\mathbf{r}_i$ and $\mathbf{g}_j$ is calculated by sliding a window of $B \times B$ pixels from the top-left corner to the bottom-right corner of the image, for a total of $M$ sliding steps:

$$l^{i,j} = \frac{1}{M} \sum_{k=1}^{M} \frac{2.\mu_k(\mathbf{r}_i).\mu_k(\mathbf{g}_j) + C_l}{\mu_k(\mathbf{r}_i)^2 + \mu_k(\mathbf{g}_j)^2 + C_l}, \quad (1)$$

where $k$ is the sliding step and $\mu_k(\cdot)$ denotes mean values of the $k^{\text{th}}$ image window. $C_l$ is a positive stabilizing constant defined as $C_l = (K_l L)^2$ where $L$ is the dynamic range of the pixel values and $K_l \ll 1$ is a small constant. Similarly, the contrast distortion between $\mathbf{r}_i$ and $\mathbf{g}_i$ is estimated using global contrast distortion in image quality (GCQ) factor defined as:

$$c^{i,j} = \frac{1}{M} \sum_{k=1}^{M} \frac{2.\sigma_k(\mathbf{r}_i).\sigma_k(\mathbf{g}_j) + C_c}{\sigma_k(\mathbf{r}_i)^2 + \sigma_k(\mathbf{g}_j)^2 + C_c}, \quad (2)$$

where $\sigma_k(\cdot)$ denotes the standard deviation of the $k^{\text{th}}$ image window. $C_c$ is a positive stabilizing constant defined as $C_c = (K_c L)^2$ where $L$ is the dynamic range of the pixel values and $K_c \ll 1$ is a small constant.

### 3) Representative Selection:
Affinity propagation (AP) [40] is applied to cluster video ROIs from the generic set defined in the normalized space defined by $\mathbf{p}_j = (\theta_i^{pitch}, \theta_i^{yaw}, \theta_i^{roll})$ and $\mathbf{u}^i = (l^i, c^i)$ measures. This clustering algorithm aims to maximize the net similarity (average distortion between ROIs and pose angles) and produce a set of exemplars. Two types of messages: *responsibility* and *availability* are exchanged between data points until a high-quality set of exemplars and corresponding clusters emerges. AP is a suitable clustering technique for DSFS because: (1) it can automatically determine the number of clusters based on the data distribution, and (2) it produced exemplars that correspond to actual ROIs. Indeed, cluster centroids produced by many prototype-based clustering methods are not necessarily actual ROIs with a real-world interpretation. Given that clustering samples simultaneously in terms of both $\mathbf{p}_j$ and $\mathbf{u}^i$ may favor certain common pose angles, a two-step clustering algorithm is proposed to preserves diversity in pose angles and illumination effects. In the first step, clustering is performed on the pose angle vector, and then the population of each pose cluster is clustered according to GLQ and GCQ metrics to find the representative luminance and contrast samples. Representative luminance and contrast samples – called *lighting exemplar* – are found along with representative pose angles – called "*pose exemplar*" (Fig. 4).

The clustering algorithm inputs a set of pose similarities $s_p(i,k) = - \parallel \mathbf{p}_i - \mathbf{p}_k \parallel^2$ indicating how well the sample $\mathbf{p}_k$ with index $k$ is similar to the sample $\mathbf{p}_i$ from the generic set. The pose responsibility $r_p(i,k)$ is defined as the accumulated evidence for how well-suited sample $\mathbf{p}_k$ is to serve as the exemplar for the sample $\mathbf{p}_i$, taking into account other potential exemplars for the sample $\mathbf{p}_i$. Evidence about whether each pose candidate exemplar would be a good exemplar is obtained from the application of the pose availability $a_p(i,k)$. The availability reflects the accumulated evidence for how appropriate it would be for sample $\mathbf{p}_i$ to choose sample $\mathbf{p}_k$ as its exemplar, taking into account the support from other samples that sample $\mathbf{p}_k$ should be an exemplar. The availabilities are initialized to zero, and the pose responsibilities are then computed iteratively using the rule of Eq.3. The availabilities are updated in each iteration using Eq.4.

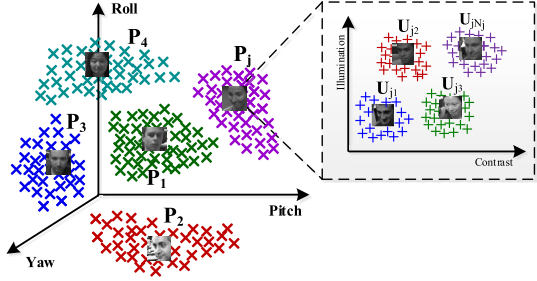

Fig. 4. An illustration of the AP clustering process.

$$r_p(i,k) = s_p(i,k) - \max_{k'|k' \neq k} \{a_p(i,k') + s_p(i,k')\}, \qquad (3)$$

$$a_p(i,k) = \min \left\{ 0, r_p(k,k) + \sum_{i'|i' \notin \{i,k\}} \max\{0, r_p(i',k)\} \right\}. \qquad (4)$$

For $\mathbf{p}_i$, the value of $\mathbf{p}_k$ that maximizes $a_p(i,k) + r_p(i,k)$ either identifies sample $\mathbf{p}_i$ as an exemplar if $k = i$, or identifies the sample that is the exemplar for the sample $\mathbf{p}_i$. The message-passing procedure is terminated after a fixed number of iterations when the local cost functions remain constant for some number of iterations. At the end of pose clustering, $K$ pose clusters $\mathbf{P} = \{\mathbf{P}_1, \mathbf{P}_2, \ldots, \mathbf{P}_j, \ldots, \mathbf{P}_K\}$ are determined, where $\mathbf{p}_i = [\theta_i^{pitch}, \theta_i^{yaw}, \theta_i^{roll}]$.

The second clustering is then applied for each pose cluster in the $l^{i,j}$ and $c^{i,j}$ measure space to find lighting exemplars. The first step computes illumination-contrast similarities $s_u(i,k) = -\| (\mathbf{u}^i - \mathbf{u}^k) \|^2$. The corresponding responsibility and availability are obtained according to:

$$r_u(i,k) = s_u(i,k) - \max_{k'|k' \neq k} \{a_u(i,k') + s_u(i,k')\}, \qquad (5)$$

$$a_u(i,k) = \min \left\{ 0, r_u(k,k) + \sum_{i'|i' \notin \{i,k\}} \max\{0, r_u(i',k)\} \right\}. \qquad (6)$$

The estimated $r_{cl}(i,k)$ and $a_{cl}(i,k)$ are combined to monitor the exemplar decisions and the algorithm is terminated when these decisions do not change for several iterations. At the end of the illumination-contrast clustering for each pose cluster $\mathbf{P}_j$, a number of $N_j$ lighting clusters $\mathbf{P}_j = \{\mathbf{U}_{j1}, \mathbf{U}_{j2}, \ldots, \mathbf{U}_{jN_j}\}$ are obtained. The central representative samples of illumination-contrast clusters in $j^{th}$ pose cluster are considered as the pose and lighting exemplars for $j^{th}$ pose as $\mathbf{u}_j^i = (l_j^i, c_j^i), 1 \leq i \leq N_j$ where $l$ and $c$ are illumination and contrast of center of $i^{th}$ illumination-contrast cluster $\mathbf{U}_{ji}$ in the $j^{th}$ pose cluster $\mathbf{P}_j$.

Larger clusters represent a greater number of generic samples, they should have more influence for the classification.

Therefore, a weight is assigned to each exemplar $\mathbf{u}_j^i$ to indicate its importance, approximated based on its cluster size, $W_{ij} = n_{ij}/n$, where $n_{ij}$ is the number of samples in the cluster $\mathbf{U}_{ij}$ and $n$ is the total number of generic samples. This selection strengthens those classes that are more representative in reconstructing a probe sample.

### B. Face Synthesis

For generating synthetic ROIs based on the representative pose and lighting conditions, 3D models of reference ROIs are reconstructed and their material-dependent layers are extracted. In the rendering process, the extracted material layers are employed as a texture of the 3D model. This model is rendered w.r.t. the pose exemplars. Following this, the illumination-dependent layers of the lighting exemplars are extracted. Finally, the lighting layers are projected on the rendered images with the same view by applying a morphing between the layers. The following subsections describe the steps proposed for the face synthesizing with DSFS.

*1) Intrinsic Image Decomposition:* Each still reference image, $\mathbf{r}_i$, is decomposed and its material-dependent layer (albedo), $\mathbf{M}_i$, based on the a texture-aware image model defined in [41]. This image decomposition method explicitly models a separate texture layer in addition to the shading layer and material layer in order to avoid ambiguity caused by textures. Explicitly modeling textures, shading layer and reflectance layer in the model depict only textureless base components, and accordingly avoid ambiguity caused by textures. Furthermore, for robustness against noise, the points are sparsely sampled for the surface normal constraint based on local variances of surface normal. This model is presented as follows:

$$\mathbf{r}_i(x,y) = \mathbf{B}^i(x,y).\mathbf{T}^i(x,y) = \mathbf{L}^i(x,y).\mathbf{M}_i(x,y).\mathbf{T}^i(x,y), \qquad (7)$$

where $\mathbf{B}(x,y) = \mathbf{L}(x,y).\mathbf{M}(x,y)$ is a base layer, and $\mathbf{L}(x,y)$, $\mathbf{M}(x,y)$ and $\mathbf{T}(x,y)$ are shading, material, and texture components at a pixel (x,y), respectively.

*2) 3D Face Reconstruction:* 3D face model of reference ROIs, $\mathbf{r}_i$, are reconstructed using the 3D Morphable Model (3DMM) technique [25], [42]. In this study, a customized version of the 3DMM is employed in which the texture fitting of the original 3DMM is replaced with image mapping. By replacing the texture fitting in the original 3DMM with 2D image mapping, an efficient method is implemented for 3D face reconstruction from one frontal face image. Basically, the shape model is defined as a convex combination of shape vectors of a set of examples in which the shape vector ($\mathbf{S}$) is defined as Eq.8 [25]. A principal components analysis is performed to estimate the statistics of the 3D shape of the faces.

$$\mathbf{S} = \bar{\mathbf{S}} + \sum_{k=1}^{m_S - 1} \alpha_k . \tilde{\mathbf{S}}_k, \qquad (8)$$

where, the 3D shape is represented by the probability distribution of faces around the averages of shape $\bar{\mathbf{S}}$ are calculated and the basis vectors $\tilde{\mathbf{S}}_j$, $1 \leq j \leq m_s$ in Eq.8 where $m_s$ is the number of the basis vectors.

Each vector $\mathbf{S}$ stores the reconstructed 3D shape in terms of x, y, z-coordinates of all vertices $\epsilon\{1, \ldots, n_s\}$ of a high-resolution 3D mesh as

$$\mathbf{S} = [X_1, Y_1, Z_1, X_2, \ldots, X_{n_s}, Y_{n_s}, Z_{n_s}]^T. \qquad (9)$$

Here, for each reference ROI, $\mathbf{r}_i$, we reconstruct the 3D shape.

$$\mathbf{S}_i = \bar{\mathbf{S}} + \sum_{j=1}^{m_S-1} \alpha_j^i \tilde{\mathbf{S}}_j, \qquad (10)$$

where $\alpha_j^i \in [0, 1]$, $1 \leq j \leq m_s$ are the shape parameters and $\mathbf{S}_i$ is the reconstructed shape of the $i^{th}$ reference still ROI $\mathbf{r}_i$. The optimization algorithm presented in [25] is employed to find optimal $\alpha_j^i$, $1 \leq j \leq m_s$, for each reference still ROI $\mathbf{r}_i$. In the next step, the extracted material layers, $\mathbf{M}_i$, are projected to the 3D geometry of the reference gallery set. Given the 3D facial shape and texture, novel poses can be rendered under various forms of the pose by adjusting the parameters of a camera model. In the rendering procedure, the 3D face is projected onto the image plane with Weak Perspective Projection which is a linear approximation of the full perspective projection.

$$\mathbf{V}^{ij} = f * \Lambda * \mathbf{R}^{ij} * (\bar{\mathbf{S}} + \sum_{j=1}^{m_S-1} \alpha_j^i) + \mathbf{t}_{2d}^i, \qquad (11)$$

where $\mathbf{V}^{ij}$ is the 2D positions of model vertexes of the $j^{th}$ reconstructed pose of $r_i$, $f$ is the scale factor, $\Lambda$ is the orthographic projection matrix $\begin{pmatrix} 1 & 0 & 0 \\ 0 & 1 & 0 \end{pmatrix}$, $\mathbf{R}^{ij}$ is the rotation matrix constructed from rotation angles *pitch*, *yaw*, *roll* and $\mathbf{t}_{2d}$ is the translation vector [43].

Since the 2D image is directly mapped to the 3D model, no corresponding color information is available for some vertices because they are occluded in the frontal face image. Consequently, it is possible that there are still some blank areas on the generated texture map. In order to correct these blank space areas, a bilinear interpolation algorithm is utilized to fill in areas of unknown texture using the known colors in the vicinity.

*3) Illumination Transferring:* For each pose exemplar, $\mathbf{p}_j$, a set of samples $\{\mathbf{I}_\mathbf{u}^{jk} \in \mathbb{R}^{d \times d} | 1 \leq k \leq N_j\}$ corresponding to the $\mathbf{u}_{jk}$, for $k = 1, 2, \ldots, N_j$, are selected as lighting exemplars. The illumination-dependent layer of each $\mathbf{I}_\mathbf{u}^{jk}$ are extracted using the same process described in section III-B1. For each pose exemplar $\mathbf{p}_j$, $N_j$ the illumination layers, $\mathbf{L}_{jk}$, for $k = 1, 2, \ldots, N_j$, are then projected to the rendered reference, $\mathbf{V}_{ij}$. This is performed by morphing between $\mathbf{L}_{jk}$ and $\mathbf{V}_{jk}$ according to the following steps:

i) detect the landmark points of $\mathbf{L}_{jk}$ and $\mathbf{V}_{ij}$ using active shape model to locate corresponding feature points. The landmark points of $\mathbf{L}_{jk}$ and $\mathbf{V}_{ij}$ are denoted as $l_{jk}$ and $v_{ij}$, respectively;

ii) define a triangular mesh over $l_{jk}$ and $v_{ij}$ via the Delaunay triangulation technique and obtain $d_l^{jk}$ and $d_v^{ij}$;

iii) coordinate transformations between $d_l^{jk}$ and $d_v^{ij}$ with affine projections on the points;

iv) warp each triangle separately from the source to destination using mesh warping technique which moves triangular patches to the newly established location to align two ROIs;

v) cross-dissolve the triangulated layers considering warped pixel locations.

In this way, a number $q = \sum_{j=1}^{K} N_j$ of synthetic ROIs are generated for each reference still ROI $\mathbf{r}_i$. Therefore, the total number of synthetic ROIs are $q_{total} = nq$. The synthetic set of ROIs for the $i^{th}$ reference still ROI are presented by $\mathbf{A}^i = [\mathbf{a}_1^i, \mathbf{a}_2^i, \ldots, \mathbf{a}_q^i] \in \mathbb{R}^{d^2 \times q}$ where $\mathbf{a}_j^i$ is the $j^{th}$ concatenated synthetic ROI for the $i^{th}$ reference still ROI. The overall process of DSFS face generation technique is formalized in Algorithm 1.

---

**Algorithm 1:** The DSFS Algorithm

**Input**: Reference set $\mathbf{R} = \{\mathbf{r}_i \in \mathbb{R}^{d \times d} | i = 1, \ldots, n\}$, and generic set $\mathbf{G} = \{\mathbf{g}_i \in \mathbb{R}^{d \times d} | i = 1, \ldots, m\}$.

1 Estimate pose angles.

2 Calculate luminance and contrast distortion measures.
`// Eq.1, Eq.2`

3 AP clustering on pose space to obtain
$\mathbf{P} = \{\mathbf{P}_1, \mathbf{P}_2, \ldots, \mathbf{P}_K\}$. `// SectionIII-A3`

4 **for** $j = 1$ *to* $K$ **do**

5     AP clustering on Illumination and contrast space for the $\mathbf{P}_j$ to obtain $\{\mathbf{u}_{ji} | 1 \leq i \leq N_j\}$.
    `// SectionIII-A3`

6 **end**

7 **for** $i = 1$ *to* $n$ **do**

8     Extract material-dependent layer of $\mathbf{r}_i$ ($\mathbf{M}_i$).
    `// SectionIII-B1`

9     Recover 3D face model of $\mathbf{r}_i$ using 3DMM ($\mathbf{S}_i$).
    `// SectionIII-B2`

10     Map the texture of $\mathbf{M}_i$ to $\mathbf{S}_i$.

11     **for** $j = 1$ *to* $K$ **do**

12         Render under $\mathbf{p}^j$ pose to obtain $\mathbf{V}_{ij}$.

13         **for** *each* $\mathbf{u}_{ji}$, $i = 1$ *to* $N_j$ **do**

14             Extract illumination-dependent layers ($\mathbf{L}_{jk}$).
            `// SectionIII-B1`

15             Morphing between $\mathbf{L}_{jk}$ and $\mathbf{V}_{ij}$ to obtain $\mathbf{A}^i$.
            `// SectionIII-B3`

16         **end**

17     **end**

18 **end**

**Output**: All sets of synthetic face ROIs under representative pose and illumination conditions.
$\mathbf{A}^i = [\mathbf{a}_1^i, \mathbf{a}_2^i, \ldots, \mathbf{a}_q^i] \in \mathbb{R}^{d^2 \times q}$, $i = 1, 2, \ldots, n$.

---

## IV. Domain-Invariant Still-to-Video Face Recognition With DSFS

In this section, a particular still-to-video FR implementation is considered (see Fig. 5) to assess the impact of using DSFS to generate synthetic ROIs to address these limitations.

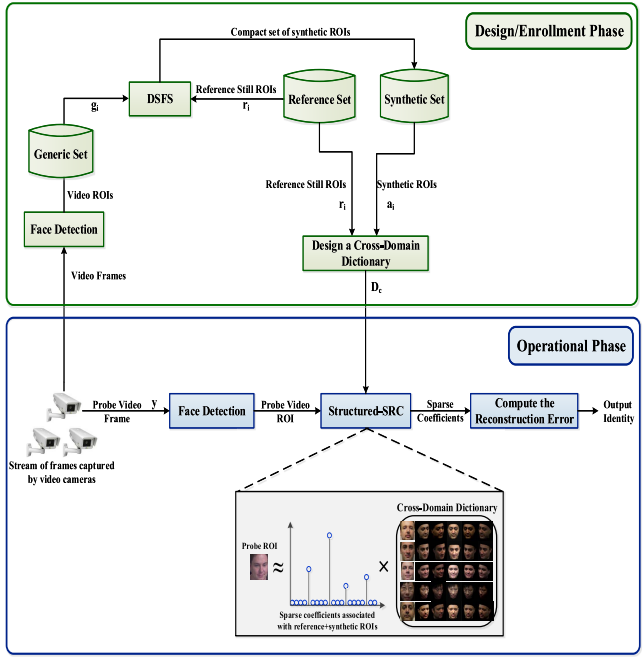

Fig. 5. Block diagram of the proposed domain-invariant SRC-based still-to-video FR system.

An augmented dictionary is constructed by employing the synthetic ROIs generated via DSFS technique, and classification is performed via a structured SRC approach. Since the synthetic ROIs for each individual (including the synthetic poses, illuminations, and etc.) form a block in this dictionary, the SRC is considered as a structured sparse recovery problem. The main steps of the proposed domain-invariant still-to-video FR with dictionary augmentation are summarized as follows:

- **Step 1**: Generation of Synthetic Facial ROIs
  In the first step, $q$ synthetic ROIs $\mathbf{A}^i = [\mathbf{a}_1^i, \ldots, \mathbf{a}_q^i] \in R^{d^2 \times q}$ are generated for each $\mathbf{r}_i$ of the reference gallery set using DSFS technique, where $q$ is the number of synthetic ROIs for each class.

- **Step 2**: Augmentation of Dictionary
  The synthetic ROIs generated through the DSFS technique are added to the reference dictionary to design a cross-domain dictionary. let $\mathbf{D}_R = [\mathbf{I_r}^1, \mathbf{I_r}^2, \ldots, \mathbf{I_r}^n] \in \mathbb{R}^{d^2 \times n}$ be the reference gallery dictionary, where $\mathbf{I_r}^i$ is the concatenated result of $\mathbf{r}_i$. The cross-domain dictionary $\mathbf{D}_C = [\mathbf{I_r}^1, \mathbf{A}^1, \ldots, \mathbf{I_r}^n, \mathbf{A}^n] \in \mathbb{R}^{d^2 \times n(q+1)}$ integrates the original and synthetic ROIs in a linear model where $\mathbf{A}^j$ is the $j^{\text{th}}$ set of synthetic ROIs added to the $j^{\text{th}}$ class. Since $q$ synthetic ROIs are added to each class, the total number of ROIs in the cross-domain dictionary are $n_c = n(q+1)$. The presented dictionary design in this work enables SRC to perform recognition with only one reference still ROI and makes it robust to the visual domain shift.

- **Step 3**: Classification
  Given a probe video ROI $\mathbf{y}$, general SRC represents $\mathbf{y}$ as a sparse linear combination of the codebook $\mathbf{D}_C$, and derives the sparse coefficients of $\mathbf{y}$ by solving the

$\ell_0$-minimization problem as follows:

$$A_{\ell_0}: \quad \min \|\mathbf{x}\|_0 \quad s.t. \quad \mathbf{y} = \mathbf{D}_C \mathbf{x}. \tag{12}$$

Since the generated synthetic ROIs for each individual form a block of the dictionary, a better classification can arise from a representation of the probe ROI produced from the minimum number of blocks from the dictionary instead of looking for the representation of a probe ROI in the dictionary of all the training data using the so-called structured SRC which its goal is to find a representation of a probe ROI that uses the minimum number of blocks from the dictionary. For a dictionary $\mathbf{D}_C = [\mathbf{D}_C[1], \mathbf{D}_C[2], \ldots, \mathbf{D}_C[n]]$ with blocks $\mathbf{D}_C[i]$, $i = 1, \ldots, n$, the block sparsity is formulated in terms of mixed $\ell_2/\ell_0$ norm as;

$$A_{\ell_2/\ell_0}: \quad \min_{\mathbf{x}} \sum_{i=1}^{n} I(\| \mathbf{x}[i] \|_2 > 0) \quad s.t. \ \mathbf{y} = \mathbf{D}_C \mathbf{x}, \tag{13}$$

where $I(.)$ is the indicator function, and $\mathbf{x}[i]$ is the $i^{th}$ block in the sparse coefficient vector $\mathbf{x}$ corresponding to the dictionary block $\mathbf{D}_C[i]$. Since each dictionary block corresponds to a specific class, $i$ represents the class index ranging from 1 to $n$ as well. This optimization problem seeks the minimum number of non-zero coefficient blocks that reconstruct the probe ROI.

Note that the optimization program $A_{\ell_2/\ell_0}$ is NP-hard since it requires searching over all possible few blocks of $\mathbf{x}$ and checking whether they span the given $\mathbf{y}$. A relaxation of this problem is obtained by replacing the $\ell_0$ with the $\ell_1$ norm and solving the Eq.14.

$$A_{\ell_2/\ell_1}: \quad \min_{\mathbf{x}} \sum_{i=1}^{n} \| \mathbf{x}[i] \|_2 \quad s.t. \ \mathbf{y} = \mathbf{D}_C \mathbf{x}. \tag{14}$$

Finally, the weighted matrix obtained in III-A3 which shows cluster weights is multiplied to the $\ell_1$-minimization term.

$$A_{\ell_2/\ell_1}: \quad \hat{x} = arg \min_{\mathbf{x}} \sum_{i=1}^{n} \|\mathbf{W}_i \ \mathbf{x}[i] \|_2 \quad s.t. \ \mathbf{y} = \mathbf{D}_C \mathbf{x}. \tag{15}$$

where

$$\mathbf{W}_i = \begin{bmatrix} w_{i1} & 0 & \cdots & 0 \\ 0 & w_{i2} & \cdots & 0 \\ \vdots & \vdots & \ddots & \vdots \\ 0 & 0 & \cdots & w_{i(q+1)} \end{bmatrix} \tag{16}$$

The class label of the probe ROIs $y$ is then determined based on the block sparse reconstruction error as follows:

$$label(\mathbf{y}) = arg \min_i \| \mathbf{y} - \mathbf{D}[i] \hat{\mathbf{x}}[i] \|_2. \tag{17}$$

In order to solve the SRC problem of equation 15, the classical alternating direction method (ADM) is considered which is an efficient first-order algorithm with global convergence [44].

- **Step 4**: Validation

  In practical FR systems, it is important to detect and then reject outlier invalid probe ROIs. We use the *sparsity concentration index (SCI)* criteria defined in [3]:

$$\text{SCI}(\hat{\mathbf{x}}) \doteq \frac{n \cdot \max_i \parallel \hat{\mathbf{x}}[i] \parallel_1 / \parallel \hat{\mathbf{x}} \parallel_1 - 1}{n - 1} \in [0, 1]. \quad (18)$$

  where $n$ is the number of classes. A probe ROI is accepted as valid if $SCI \geq \tau$ and otherwise rejected as invalid, where $\tau \in (0, 1)$ is a threshold.

The still-to-video face recognition process through dictionary augmentation is formalized in Algorithm 2.

---

**Algorithm 2:** A SRC-Based Still-to-Video FR System

---

**Input**: Reference face models of $n$ classes enlisted in the gallery $\mathbf{R} = \{\mathbf{r}_i \in \mathbb{R}^{d \times d} | i = 1, \ldots, n\}$, generic set $\mathbf{G} = \{\mathbf{g}_i \in \mathbb{R}^{d \times d} | i = 1, \ldots, m\}$, threshold $\tau$, and a probe ROI **y**.

1 Generate $nq$ synthetic ROIs for each class using the DSFS method.

2 Build the cross-domain dictionary $\mathbf{D}_C$ by adding the synthetic ROIs to the reference gallery set.

3 Solve the $A_{\ell_2/\ell_1}$ problem using ADM technique.
   `// Eq. 15`

4 **if** $SCI \geq \tau$ **then**                    `// Eq. 18`

5 | Find the class label **y**.              `// Eq. 17`

6 **else**

7 | Reject as invalid.

**Output**: Class label of **y**.

---

## V. EXPERIMENTAL METHODOLOGY

### A. Databases

In order to validate the proposed DSFS for still-to-video FR under real-world surveillance conditions, extensive experiments were conducted on two publicly available datasets – COX-S2V [45] and Chokepoint [46]. These datasets were selected because they are the most representative for watch-list screening applications. They contain a high-quality reference image per subject captured under controlled condition (with a still camera), and lower-quality surveillance videos for each subject captured under uncontrolled conditions (with surveillance cameras).

COX-S2V dataset [45] contains 1000 individuals, with 1 high-quality still image and 4 low-resolution video sequences per individual simulating video surveillance scenario. In each video, an individual walk through a designed S-shape route with changes in illumination, scale, and pose. The Chokepoint dataset [46] consists of 25 individuals walking trough portal 1, and 29 individuals walking trough portal 2. The recording of portal 1 and portal 2 are one month apart. A camera rig with 3 cameras is used for simultaneously recording the entry of a person during four sessions with changes in illumination conditions, pose, and misalignment. In total, the dataset consists of 54 video sequences and 64,204 face images.

### B. Experimental Protocol

With the Chokepoint database, 5 individuals are randomly chosen as watch-list individuals that each individual includes a high-quality frontal captured image. Prior to each experiment, the video data is split into 3 parts. ROIs are extracted from the video sequences of 10 other individuals selected at random as a generic set to represent capture conditions. ROIs of the video sequences of the remaining individuals along with video sequences of the 5 already selected watch-list individuals are employed for testing. In order to obtain representative results, this process is repeated 5 times with a different random selection of watch-list and generic set individuals, and the average accuracy is reported with mean and standard deviation over all the runs. With COX-S2V, 30 individuals are randomly considered as watch-list individuals including a high-quality captured image per each individual. Their corresponding low-quality video sequences along with ROIs of the video sequences of 100 other individuals are employed for testing. The ROIs extracted from the video sequences of 100 other individuals are selected at random as a generic set to represent capture conditions. This process is replicated 5 times with different stills and videos of watch-list individuals, and the average accuracy is reported with mean and standard deviation over all the runs. During the enrollment, the ROIs of the generic set of faces captured from video trajectories across all ODs (i.e., global modeling) are extracted using the Viola-Jones face detection algorithm [47]. Face detection is also applied to still images prior to face synthesis. An estimation of luminance and contrast are measured for each video ROIs in the generic set where the constant values of $K_l$ and $K_C$ are set to 0.01 and 0.03, respectively as proposed in [39]. Pose angles are estimated. Then, AP clustering is applied to the generic set, where $q$ representative video ROIs are selected under various pose, illumination and contrast conditions, and a weight is assigned to each exemplar according to the cluster size. Then, $q$ synthetic face images are generated for each individual based on the information obtained from these selected exemplars. Recall that AP clustering seeks exemplars (samples that are representative of clusters), and automatically determines $k$ and $q$, the number of clusters, for each independent replication. The cross-domain dictionary is then designed using the reference still and synthetic ROIs. During the operational phase, recognition is performed by coding the probe image over the cross-domain dictionary regarding the weights obtained in the enrollment domain. Throughout the experiments, the sparsity parameter $\Lambda$ is fixed to 0.005. For reference, the still-to-video FR system based on individual-specific SVMs is also evaluated. During the enrollment, a non-linear SVM classifier with RBF kernel is trained for each individual using target ROIs (reference still of the individual plus the related synthetically face images) versus non-target ROIs (reference still of cohort persons plus their synthetic face images).

### C. Performance Measures

To assess the ability of face synthesizing techniques to address shifts between OD and ED, a domain shift quantification (DSQ) measure is employed. With this measure, the similarity between a dictionary designed using synthetic ROIs ($\mathbf{D}_A$) is compared with a dictionary formed with images collected from the OD ($\mathbf{D}_R$) by measuring the mean pixel error between the dictionaries. Given two dictionaries $\mathbf{D}_A$ and $\mathbf{D}_R$ with the same number of images, the DSQ measure is defined as $Q_{dsq} = \|\mathbf{D}_R^T\mathbf{D}_A\|_F$ where a higher value indicates less domain shift [37]. The accuracy of the still-to-video FR system is assessed per individual of interest at the transaction level, using the receiver operating characteristic (ROC) space, where the true positive rates (TPRs) are plotted as a function of false positive rates (FPRs) over all threshold values. TPR is the proportion of target ROIs that correctly classified as individuals of interest over the total number of target ROIs, while FPR is the proportion of non-target ROIs incorrectly classified as individuals of interest over the total number of non-target ROIs. The area under ROC curve is a global scalar measure of accuracy that can be interpreted as the probability of correct classification over the range of TPR and FPR. Accordingly, accuracy of FR systems is estimated using the partial area under ROC curve pAUC(10%) (using the AUC at 0 <FPR≤ 0.1%). Since the number of target and non-target data are imbalanced, the area under precision-recall curves (AUPR) is also used to estimate the performance of FR systems.

## VI. RESULTS AND DISCUSSION

This section first presents some examples of synthetic faces generated using the DSFS technique and compares them with synthetic faces generated using state-of-the-art face synthesizing methods: 3DMM [25], and 3DMM-CNN [26]. Then, the performance of still-to-video FR systems based on individual-specific SVMs and on SRC is presented when using these synthetic facial ROIs for system design. FR performance is assessed when increasing the number of synthetic ROIs per each individual according to pose angles and lighting effects. To characterize the impact on performance, these systems are tested with a growing number of synthetic ROIs and generic training set, and compared with several relevant state-of-the-art still-to-video FR systems: ESRC [8], RADL [19], SVDL [17], LGR [20], and Flow-based face frontalization [4]. The final experiment compares the performance of a system designed with synthetic ROIs obtained with DSFS, to a system designed with a growing number of randomly-selected synthetic ROIs. The dataset,face synthesizing and face recognition experiments can be viewed at https://github.com/faniamokhayeri/DSFS.

### A. Face Synthesis

This subsection presents examples of pose and lighting exemplars obtained by clustering of facial trajectories in the captured condition space. Figs. 6a and 6b show an example of pose clusters obtained with Chokepoint video trajectories of 10 individuals, and with COX-S2V video trajectories of 100 individuals. In this experiment, $k_1 = 9$ and $k_2 = 7$ pose clusters (exemplars) are typically determined with the Chokepoint and COX-S2V videos, respectively. The second level of clustering is then applied in the illumination and contrast measure space on each pose clusters. Figs. 6c and 6d show the exemplars selected based on both pose and lighting with the proposed representative selection of DSFS (see section III-A). Overall, $q_1 = 22$ and $q_2 = 18$ exemplars were typically selected based on both pose and lighting clusters determined in Chokepoint and COX-S2V videos, respectively.

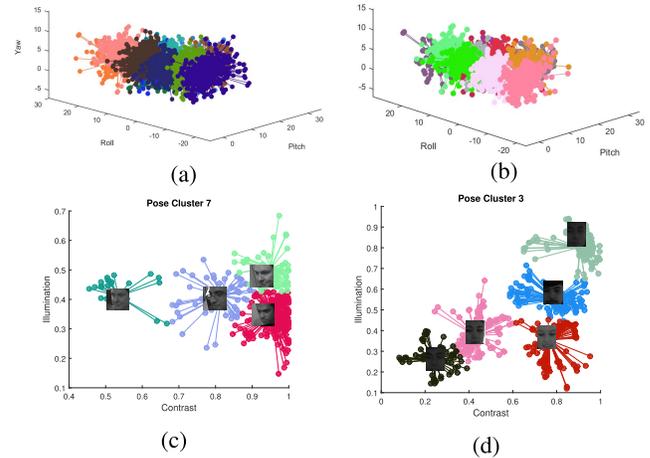

Fig. 6. Examples of representative selection results using AP clustering technique in terms of (a, b) pose angles, and (c, d) luminance and contrast with Chokepoint dataset on video sequences of 10 individuals and COX-S2V dataset on video sequences of 100 individuals, respectively, where the center of clusters show exemplars.

Fig. 7 show examples of synthetic ROIs generated under different pose, illumination and contrast conditions using the DSFS technique on the Chokepoint (Figs. 7a and 7b) and COX-S2V datasets (Figs. 7c and 7d), where Basel Face Model are used as generative 3D shape model [42]. In Fig.8, the quality of synthetic faces generated under different pose via DSFS, 3DMM [25], and 3DMM-CNN [26] techniques are compared. The synthetic ROIs generated using the DSFS are also evaluated quantitatively. Table. I shows the DSQ values of the DSFS and other state-of-the-art face synthesizing methods including 3DMM [25] 3DMM-CNN [26], and SHBMM [27] on Chokepoint and COX-S2V datasets. Higher DSQ values indicate a smaller domain shift, and potentially higher recognition rate between the corresponding two domains. The results are provided under the two following scenarios.

*1) Frontal View:* In the first experiment, 5 individuals in ED are randomly selected. A set of synthetic face are generated with a frontal view under various lighting effects from the still ROI of each individual to design $\mathbf{D}_A$. The corresponding video ROIs in the OD under the frontal view are then collected to form $\mathbf{D}_R$. Finally, the DSQ is measured for the $\mathbf{D}_A$ and $\mathbf{D}_r$ dictionaries.

*2) Profile View:* In the second experiment, 5 individuals in ED are again randomly selected. Their synthetic ROIs are generated with profile view and different illumination conditions to form $\mathbf{D}_A$. The corresponding video ROIs in OD under profile view are collected to construct $\mathbf{D}_R$. Finally, the DSQ is estimated for the dictionaries.

As shown in Table. I, DSQ values of DSFS method are higher followed most closely by SHBMM in both scenarios. Accordingly, the cross-domain dictionary designed by the synthetic ROIs generated via the DSFS method is most suitable to reduce visual domain shifts and potentially achieve a higher level of accuracy. These results are in line with the recognition performance results.

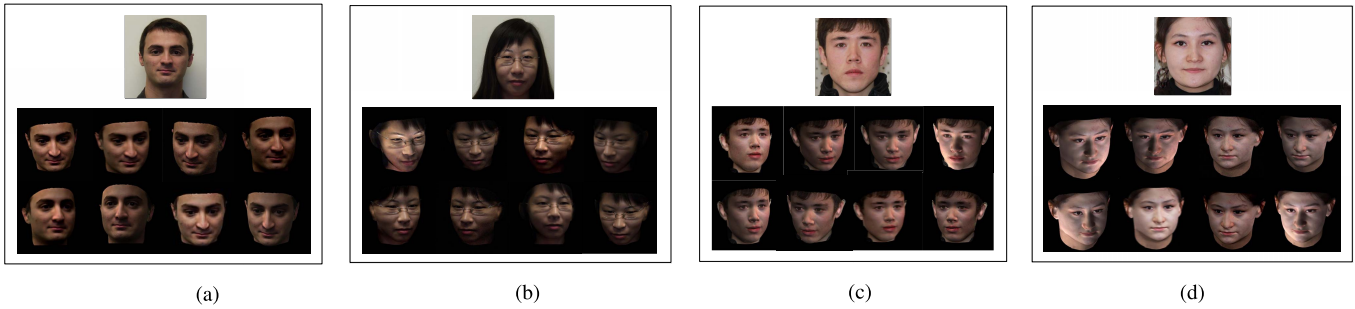

Fig. 7. Examples of synthetic ROIs generated under different capture conditions using the DSFS technique with Chokepoint (a, b) and COX-S2V (c, d) datasets.

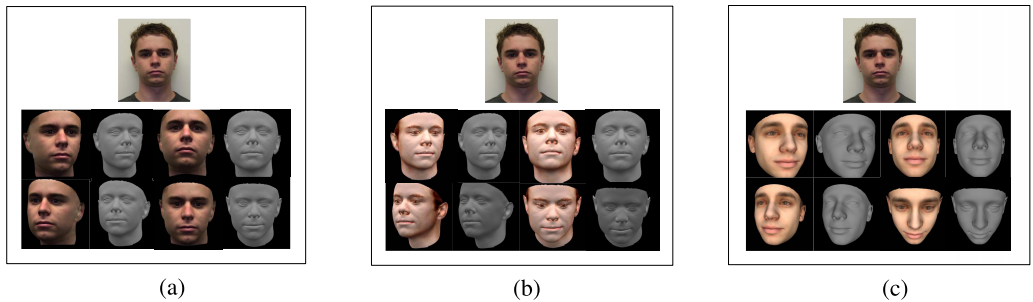

Fig. 8. Synthetic face images generated under different pose via (a) DSFS, (b) 3DMM, (c) 3DMM-CNN with Chockpoint dataset.

TABLE I
AVERAGE DSQ VALUE FOR FRONTAL AND PROFILE
VIEWS ON CHOKEPOINT AND COX-S2V DATASETS

| Technique | Chokepoint database | | COX-S2V database | |
|---|---|---|---|---|
| | Frontal View | Profile View | Frontal View | Profile View |
| 3DMM[25] | 8.27 | 7.38 | 8.24 | 7.63 |
| 3DMM-CNN[26] | 7.61 | 7.03 | 7.57 | 7.29 |
| SHBMM[27] | 9.16 | 7.26 | 9.34 | 7.71 |
| **Proposed DSFS** | **9.53** | **8.17** | **9.58** | **8.39** |

### B. Face Recognition

In this subsection, the performance achieved using the still-to-video FR system based on SRC and DSFS (see Section IV) is assessed experimentally. For reference, the still-to-video FR system based on individual-specific SVMs is also evaluated.

*1) Pose Variations:* The still-to-video FR system is evaluated versus the number of synthetic ROIs that incorporate growing facial pose. Figs. 9a and 9d show the average AUC and AUPR obtained by increasing the number of synthetic ROIs generated using DSFS from $k$ representative pose angles ($\mathbf{p}_i$, $i = 1 \ldots k$) and with fixed lighting condition. Results indicate that by adding extra synthetic ROIs generated under representative pose angles allows to outperform baseline systems designed with an original reference still ROI alone. AUC and AUPR accuracy increases by about 10%, typically with only $k_1 = 9$ and $k_2 = 7$ synthetic pose ROIs for Chokepoint and COX-S2V datasets, respectively.

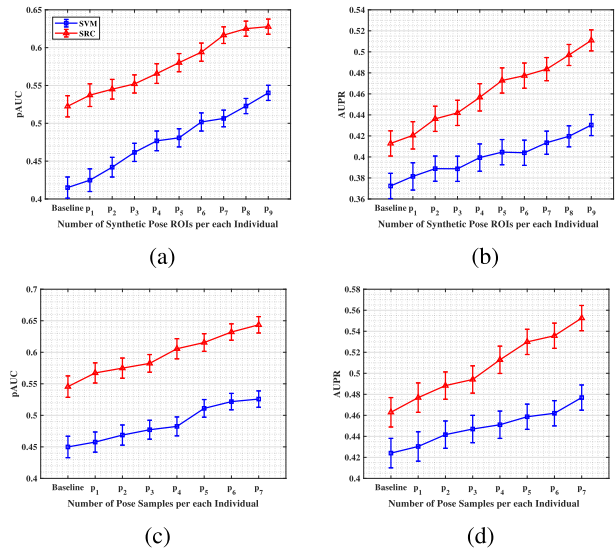

Fig. 9. Average AUC and AUPR versus the number of synthetic ROIs generated with DSFS according to various pose and fixed illumination. The still-to-video FR system employs either SVM and SRC classifiers on Chokepoint (a, b) and COX-S2V (c, d) databases.

*2) Mixed Pose and Illumination Variations:* The performance of still-to-video FR systems is assessed versus the number of synthetic ROIs generated under both pose and lighting effects. Figs. 10a and 10d show average AUC and AUPR obtained by increasing the number of synthetic ROIs used to design SRC and SVM classifiers on the Chokepoint and COX-S2V databases, where $\mathbf{A}_i$ is a set of synthetic ROIs generated using DSFS technique under various pose and illumination conditions. Adding synthetic ROIs generated under various pose, illumination and contrast conditions allows to significantly outperform the baseline system designed with the original reference still ROI alone. AUC and AUPR accuracy increases by about 40%, typically with only $q_1 = 24$ and $q_2 = 18$ synthetic ROIs for Chokepoint and COX-S2V datasets, respectively. As shown in Figs. 10a and 10d, accuracy for DSFS+SRC trends to stabilize to its maximum value when the size of the generic set is greater than $q$ in DSFS. To view performance stabilizing with more than $q$ synthetic ROIs, additional samples were selected randomly among AP clusters. Note that the Chokepoint dataset contains faces captured for a range illumination conditions with various densities. Hence, some exemplars may represent many video ROIs. Our method assigns higher weights to such distributions and may yield a higher level of performance. The results obtained with DSFS are also compared to still-to-video FR systems that exploit state-of-the-art face synthesis techniques including 3DMM [25] (with randomly selected images), and SHBMM [27] (with 9 spherical harmonic basis images). As shown in Fig. 10, DSFS always outperforms these other techniques.

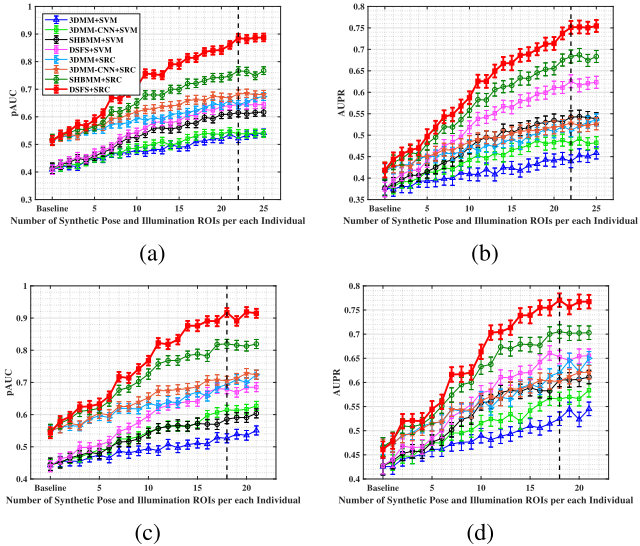

Fig. 10. Average AUC and AUPR versus the number of synthetic ROIs generated with DSFS, 3DMM, and SHBMM according to pose and lighting effects where still-to-video FR system employs either SVM and SRC classifiers on Chokepoint (a, b) and COX-S2V (c, d) databases.

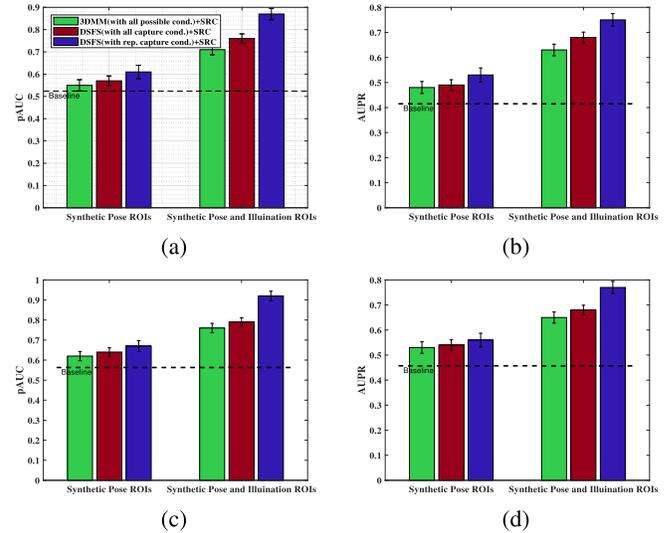

Fig. 11. Average AUC and AUPR of a still-to-video FR system designed with representative synthetic ROIs vs a system designed with randomly generated synthetic ROIs on Chokepoint (a, b) and COX-S2V (c, d) datasets.

*3) Impact of Representative Selection:* Without prior knowledge of the OD, synthetic faces are generated according to a uniform distribution. Adding a large number of synthetic ROIs to the dictionary as needed to cover all possible cases can significantly increase the time and memory complexity of FR systems and, more importantly, may cause over-fitting. The proposed DSFS technique extracts representative information from the OD to produce a compact set of synthetic ROIs that are robust to intra-class variations in the OD. In order to evaluate the impact of the synthetic ROIs generated based on representative information (i.e., pose and lighting cluster instances), 3 dictionaries are designed for SRC: (1) a dictionary designed with representative synthetic ROIs (DSFS technique); (2) a dictionary designed with the synthetic ROIs under all capture conditions (DSFS without AP clustering); and (3) a dictionary designed under all possible conditions.

The first scenario evaluates the impact of representative selection in terms of *pose* with 3 dictionaries. The first dictionary typically employs $k_1 = 9$ and $k_2 = 7$ representative synthetic pose ROIs generated with the DSFS technique for Chokepoint and COX-S2V datasets, respectively. The second dictionary employs 100 synthetic pose ROIs generated by DSFS technique under all OD capture conditions. The third dictionary employs 180 synthetic pose ROIs generated by 3DMM in a set of rotation angles ranging from to $-60$ to $+60$ (Figs. 11a and 11d). The second scenario evaluates the impact of representative selection in terms of both *pose and illumination conditions* with 3 dictionaries designed for SRC. The first dictionary employs $q_1 = 22$ and $q_2 = 18$ representative synthetic ROIs generated under different pose and illumination with the DSFS technique for Chokepoint and COX-S2V datasets, respectively. The second dictionary employs 100 synthetic ROIs generated under different pose and illumination by DSFS technique under all OD capture conditions. The third dictionary employs 180 synthetic pose ROIs generated under different pose and illumination by 3DMM (Figs. 11a and 11d). The results in Fig. 11 suggest that augmenting the dictionary using representative synthetic ROIs with the DSFS technique yields a higher level of accuracy, particularly under both pose and illumination conditions.

The impact of the proposed representative selection technique is also assessed based on the various pose estimation methods including DRMF [38], ERT [48], and OpenFace [49]. For this, the performance of the FR system are compared according to the different pose estimation techniques under combined variations of identity, pose, and illumination conditions. In this experiment, the 5 individuals of Chokepoint database and 30 individuals of COX-S2V database are used. With Chokpoint dataset, $q_1 = 22$, $q_2 = 20$, $q_3 = 24$ representative samples with DRMF, ERT, and OpenFace are obtained, respectively, and With COX-S2V dataset, $q_1 = 18$, $q_2 = 16$, $q_3 = 21$ representative samples with DRMF, ERT, and OpenFace are obtained, respectively. Figs.12 show the average AUC and AUPR obtained by increasing the number of synthetic ROIs generated from $q_1$, $q_2$, and $q_3$ representative pose angles obtained with different pose estimation techniques. Then, the error of each pose estimation technique is computed based on the normalized distance of each landmark to its ground truth position (see Table. II). It can be observed from the results that when error of pose estimation is low, the accuracy increases. The results suggest that the robustness of pose estimation techniques to nuisance factors has an impact on the performance of the FR system.

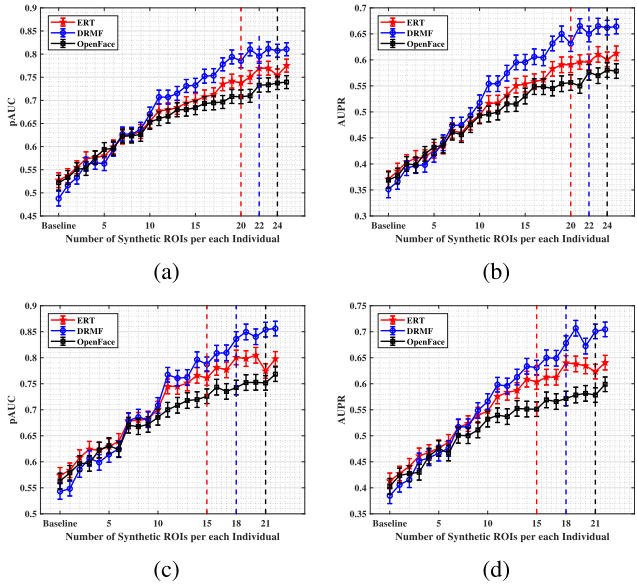

Fig. 12. Average AUC and AUPR accuracy obtained by increasing the number of the synthetic ROIs generated using DSFS from different representative pose angles (obtained with different pose estimation techniques) on Chokepoint (a, b) and COX-S2V (c, d) datasets.

TABLE II
AVERAGE ERROR RATE OF POSE ESTIMATION FOR FRONTAL AND PROFILE VIEWS ON CHOKEPOINT AND COX-S2V DATASETS

| | Error rate of pose estimation | | | |
|---|---|---|---|---|
| | Chokepoint database | | COX-S2V database | |
| Technique | Frontal View | Profile View | Frontal View | Profile View |
| DRMF[38] | 0.35 | 0.46 | 0.31 | 0.41 |
| ERT[48] | 0.39 | 0.52 | 0.35 | 0.46 |
| OpenFace[49] | 0.42 | 0.55 | 0.38 | 0.48 |

The impact of illumination transferring on the DSFS technique is further evaluated. For this, the DSQ value (V-C) of the DSFS technique is compared based on the proposed illumination transferring method and the method presented in [50] that transfer illumination through adaptive layer decomposition. In this experiment, we consider 5 and 30 individuals of Chokepoint and COX-S2V databases, respectively. With the proposed technique, DSQ of the Chokepoint and COX dataset are $DSQ = 8.63$ and $DSQ = 8.81$, respectively. With the adaptive layer decomposition method, DSQ of the Chokpoint and COX dataset are $DSQ = 7.24$ and $DSQ = 7.39$, respectively. It can be concluded that the robustness of illumination transferring to unrelated distortions has an impact on the performance of DSFS technique. Since the shading decomposition technique employed in our illumination transferring technique [41] is able to explicitly model the texture layer, the decomposed shading layer does not have any textures. As a result, It can avoid ambiguity caused by textures. However, weighted least squares filter employed in [50] cannot deal with nuisance factors.

### C. Comparison With Reference Techniques

With the above experimental setting, we compare the recognition rate of the DSFS technique with the recent face synthesizing methods (3DMM [25], SHBMM [27]) in a still-to-video FR framework. We also present the impact of using face synthesizing along with KSVD dictionary learning [51].

Following this, the recognition rate of the DSFS technique with existing generic learning techniques (ESRC [8], RADL [19], SVDL [17], LGR [20]) is compared that regularization parameter $\Lambda$ is set to 0.005. Note that the performance of the face synthesizing techniques is evaluated w/o dictionary learning. We also compared the DSFS results with the results obtained by Flow-based face frontalization method [4]. Table. III lists and compares the recognition performance where the results (recognition rate) are illustrated by the mean and standard deviation of 5 runs.

*1) Generic Set Dimension:* In this subsection, the results of DSFS technique and some generic learning techniques are evaluated based on the size of the generic set. Given $N$ generic images in the operational domain, the recognition rate of the approaches is compared with increasing value of $N$. In this comparison, each system is considered as a black box, and their recognition rate is shown for a range of different numbers of inputs. Figs. 13a and 13b show that for many generic learning techniques, intra-class variation of a small number of individuals in operational environment is sufficient to largely improve the recognition rate. In particular, it can be observed from Figs.13 that when more generic images are used, the accuracy increases significantly from our method and RADL [19] technique, while the accuracies of other state-of-the-art methods do not change significantly. This shows that the proposed representative selection method is able to adequately select the representative faces out of a large set of faces.

Next, we compare the computational complexity in terms of average running time for each individual as well as number of inner products needed per each iteration. Figs. 14a and 14b shows the computational complexity in terms of number of inner products with a growing number of synthetic ROIs.

Table. IV compares the complexity of the proposed DSFS-SRC algorithm with RADL [19], LGR [20], and flow-based frontalization techniques on Chokepoint and COX-S2V

TABLE III

COMPARATIVE TRANSACTION LEVEL ANALYSIS OF THE PROPOSED FR APPROACH AND RELATED STATE-OF-THE ART
FR METHODS WITH CHOKEPOINT AND COX-S2V DATABASES

| Category | Technique | Classifier | Chokepoint database | | COX-S2V database | |
|---|---|---|---|---|---|---|
| | | | pAUC | AUPR | pAUC | AUPR |
| | Baseline | SRC | 0.516±0.033 | 0.415±0.035 | 0.548±0.031 | 0.457±0.036 |
| **Generic Learning** | ESRC[8] | SRC<br>SRC-KSVD | 0.798±0.029<br>0.809±0.024 | 0.651±0.032<br>0.672±0.022 | 0.827±0.028<br>0.831±0.018 | 0.695±0.032<br>0.715±0.020 |
| | RADL[19] | SRC | 0.847±0.025 | 0.724±0.031 | 0.883±0.024 | 0.753±0.027 |
| | LGR[20] | SRC | 0.841±0.028 | 0.717±0.024 | 0.877±0.026 | 0.744±0.025 |
| | SVDL[17] | SRC | 0.823±0.021 | 0.703±0.029 | 0.839±0.022 | 0.724±0.031 |
| **Face Synthesizing** | 3DMM[25] | SRC<br>SRC-KSVD | 0.663±0.035<br>0.712±0.032 | 0.523±0.037<br>0.605±0.032 | 0.702±0.031<br>0.732±0.031 | 0.562±0.032<br>0.641±0.028 |
| | 3DMM-CNN[26] | SRC<br>SRC-KSVD | 0.672±0.025<br>0.716±0.024 | 0.516±0.026<br>0.585±0.025 | 0.705±0.025<br>0.741±0.026 | 0.552±0.025<br>0.603±0.025 |
| | SHBMM[27] | SRC<br>SRC-KSVD | 0.721±0.032<br>0.773±0.026 | 0.593±0.040<br>0.671±0.022 | 0.735±0.032<br>0.784±0.027 | 0.607±0.041<br>0.681±0.028 |
| **Face Frontalization** | Flow-based[4] | SRC | 0.822±0.021 | 0.711±0.024 | 0.843±0.022 | 0.719±0.023 |
| **Face Synthesizing + Generic Learning** | Proposed DSFS | SRC | **0.897±0.023** | **0.751±0.027** | **0.916±0.18** | **0.775±0.25** |

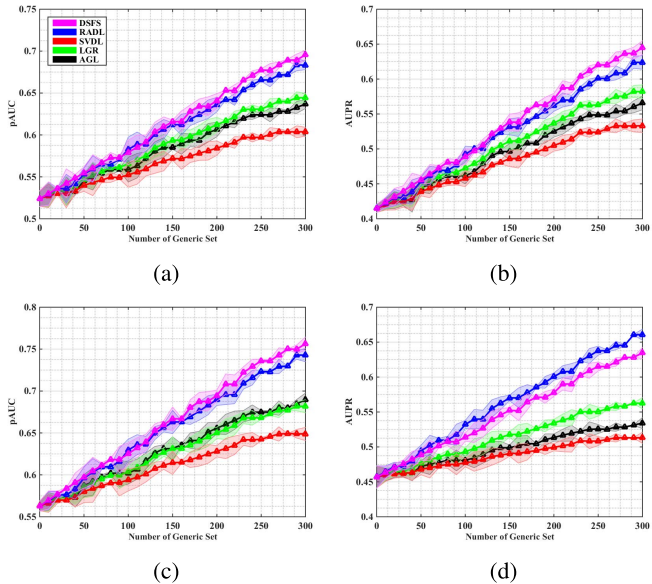

Fig. 13. Average AUC and AUPR accuracy obtained by increasing the size of the generic set in the (synthetic) variant dictionary on Chokepoint (a, b) and COX-S2V (c, d) datasets.

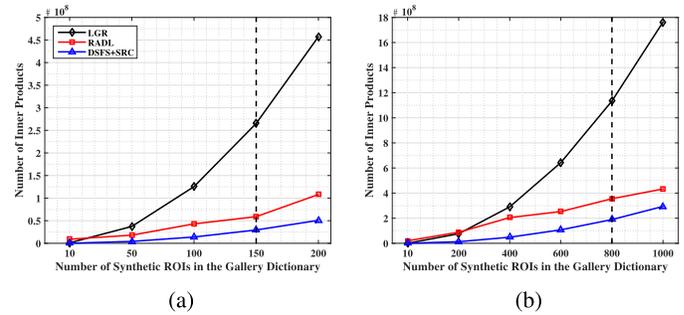

Fig. 14. Time complexity versus the number of synthetic ROIs on Chokepoint (a) and COX-S2V (b) data.

TABLE IV

AVERAGE COMPUTATIONAL COMPLEXITY OF THE
DSFS AND STATE-OF-THE-ART METHODS ON
CHOKEPOINT AND COX DATASETS

| Technique | Chokepoint database | | COX-S2V database | |
|---|---|---|---|---|
| | No.inner products | Run time(s) | No.inner products | Run time(s) |
| RADL[19] | 52,650,000 | 3.12 | 350,500,000 | 6.17 |
| LGR[20] | 273,100,000 | 7.13 | 1,147,200,000 | 11.35 |
| Face Frontalization+SRC[4] | 40,510,000 | 3.14 | 245,320,000 | 4.08 |
| **Proposed DSFS+SRC** | **32,450,000** | **1.55** | **190,250,000** | **2.61** |

datasets per each iteration. The experiments are conducted in MATLAB R2016b (64bit) Linux version on a PC workstation with an INTEL CPU (3.41-GHz) and 16GB RAM.

The results show our proposed method that is a joint use of generic learning and face synthesizing achieves superior recognition results compared to the other methods under the same configuration which verifies that our face synthesizing technique better preserves identity information. Although RADL, LGR, and the flow-based face frontalization techniques can achieve comparable accuracy to our approach, they are computationally expensive. It can be concluded that augmenting the SRC with synthetic ROIs generated by DSFS technique has a good recognition rate with less computational cost than other state-of-the-art methods. The main reason is that the dictionary designed by DSFS technique is able to represent real-world capture conditions and does not require any traditional dictionary learning process.

## VII. Conclusions

This paper proposes a domain-specific face synthesizing (DSFS) technique to improve the performance of still-to-video FR systems when surveillance videos are captured under various uncontrolled conditions, and individuals are recognized based on a single facial image. The proposed approach takes advantage of operational domain information from the generic set that can effectively represent probe ROIs. A compact set of synthetic faces is generated that resemble individuals of interest under capture conditions relevant to the operational domain. For proof-of-concept validation, an augmented dictionary with a block structure is designed based on DSFS, and face classification is performed within a SRC framework. Our experiments on the Chokepoint and COX-S2V datasets show that augmenting the reference discretionary of still-to-video FR systems using the proposed DSFS approach can provide a higher level of accuracy compared to state-of-the-art approaches, with only a moderate increase in its computational complexity. The results indicated that face synthesis alone (without recovering the OD information) cannot effectively resolve the challenges of the SSPP and visual domain shift problems. With DSFS, generic learning and face synthesis operate complementarily. The proposed DSFS technique could be improved to generate synthetic faces with expression variations for a robust FR. In addition, to improve performance, the representative synthetic ROIs generated using DSFS could be applied to generate local camera-specific ROIs. DSFS is general in that synthetic ROIs could be applied to train or fine-tune a multitude of face recognition systems like deep CNNs, with information that robust models to specific operational domains.

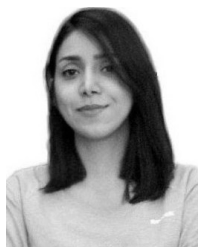

**Fania Mokhayeri** is currently pursuing the Ph.D. degree in computer science with École de technologie supérieure, Université du Québec, Montréal, Canada. Her research interests include computer vision, machine learning, face recognition, deep learning, and video surveillance applications.

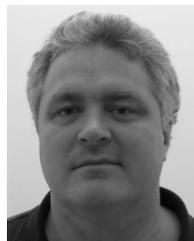

**Eric Granger** received the Ph.D. degree in EE from École Polytechnique de Montréal in 2001. He was a Defense Scientist with DRDC, Ottawa, from 1999 to 2001, and in R&D with Mitel Networks from 2001 to 2004. He joined École de technologie supérieure, Université du Québec, Montreal, Canada, in 2004, where he is currently a Full Professor and the Director of LIVIA, a research laboratory focused on computer vision and artificial intelligence. His research interests include adaptive pattern recognition, machine learning, computer vision, and computational intelligence, with applications in biometrics, face recognition and analysis, video surveillance, and computer/network security.

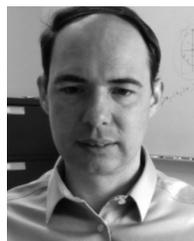

**Guillaume-Alexandre Bilodeau** received the B.Sc.A. degree in computer engineering and the Ph.D. degree in electrical engineering from the Université Laval, Canada, in 1997 and 2004, respectively. He was appointed as an Assistant Professor with the Polytechnique Montréal, Canada, in 2004, where he was an Associate Professor in 2011. Since 2014, he has been a Full Professor with Polytechnique Montréal. His research interests encompass image and video processing, video surveillance, object recognition, content-based image retrieval, and medical applications of computer vision. He is a member of the Province of Québec's Association of Professional Engineers and the REPARTI research network.